\documentclass{article}

\usepackage[final,nonatbib]{neurips_2019}

\usepackage{enumitem}
\usepackage[utf8]{inputenc} 
\usepackage[T1]{fontenc}    
\usepackage{hyperref}       
\usepackage{url}            
\usepackage{booktabs}       
\usepackage{amsfonts}       
\usepackage{nicefrac}       
\usepackage{microtype}      
\usepackage{color}
\usepackage{geometry}
\usepackage[tight]{bibhacks}

\usepackage{amsmath}
\usepackage{amssymb}
\usepackage{graphicx}
\usepackage[colorinlistoftodos]{todonotes}
\usepackage{changes}
\usepackage{listings}
\usepackage{lipsum}
\usepackage{wrapfig}

\title{Intrinsic dimension of data representations in deep neural networks}

\author{%
  Alessio Ansuini\\
  International School for Advanced Studies\\ 
  \texttt{alessioansuini@gmail.com}
  \And
  Alessandro Laio\\
  International School for Advanced Studies\\ 
  \texttt{laio@sissa.it}
  \AND
  Jakob H. Macke\\
  Technical University of Munich\\
  \texttt{macke@tum.de}
  \And
  Davide Zoccolan\\
  International School for Advanced Studies\\ 
  \texttt{zoccolan@sissa.it}
}

\begin{document}

\maketitle

\begin{abstract}
Deep  neural networks  progressively transform their inputs across multiple processing layers. What are the geometrical properties of the representations learned by these networks?
Here we study the intrinsic dimensionality (ID) of data-representations, i.e. the minimal number of parameters needed to describe a representation.
We find that, in a trained network, the ID is orders of magnitude smaller than the number of units in each layer. Across layers, the ID first increases and then progressively decreases in the final layers. Remarkably, the ID of the last hidden layer predicts classification accuracy on the test set. These results can  neither be found by linear dimensionality estimates  (e.g., with principal component analysis), nor in representations that had been artificially linearized. They are neither found in untrained networks, nor in networks that are trained on randomized labels. This suggests that neural networks  that can generalize are those that transform the data into low-dimensional, but not necessarily flat
manifolds.
\end{abstract}

\section{Introduction}

Deep neural networks (DNNs), including convolutional neural networks (CNNs) for image data, are among the most powerful tools for supervised data classification. 
In DNNs, inputs are sequentially processed across multiple layers, each performing a nonlinear transformation from a high-dimensional vector to another high-dimensional vector. Despite the empirical success and widespread use of DNNs, we still have an incomplete understanding about why and when they work so well -- in particular, it is not clear yet why they are able to generalize well to unseen data, not withstanding their massive overparametrization \cite{Zhang2017}. While progress has been made  recently [e.g.\ \cite{neyshabur2018towards,lampinen2018analytic}], guidelines for selecting architectures and training procedures are still largely based on heuristics and domain knowledge.

A fundamental geometric property of a data representation in a neural network is its \emph{intrinsic dimension} (ID), i.e., the minimal number of coordinates which are necessary to describe its points without significant information loss. 
It is widely appreciated that deep neural networks are over-parametrized, and that there is substantial redundancy amongst the weights and activations of deep nets -- e.g., several studies in network compression have shown that many weights in deep neural networks can be pruned without significant loss in classification performance \cite{denil2013predicting,LeCunDenker_90}.
Linear estimates of the ID in DNNs have been computed theoretically and numerically in simplified models \cite{huang2018mechanisms}, and local estimates of the ID developed in \cite{amsaleg2015estimating} have been related to robustness properties of deep networks to adversarial attacks \cite{amsaleg2017vulnerability,Xingjun2018}, showing that a low local intrinsic dimension correlates positively with robustness. 
Local ID of object manifolds can also be estimated at several locations on the tangent space, and was found to decrease along the last hidden layers of AlexNet \cite{yu2018curvature,krizhevsky2012imagenet}.  Both linear and nonlinear dimensionality reduction techniques have been used extensively to visualize computations in deep networks \cite{raghu2017svcca,morcos2018insights,BarrettMorcos_19}. Estimates of the  local ID \cite{amsaleg2015estimating} can assume very low values in the last hidden layers \cite{Ma_2018}, and can be used to signal the onset of overfitting in  the presence of noisy labels. Thus, local ID can be used to drive learning towards high generalization solutions, also in conditions of severe noise contamination, showing a connection between intrinsic dimension and generalization during training of specific models \cite{Ma_2018}.

\begin{wrapfigure}{r}{.55\textwidth}
\includegraphics[width=.7\textwidth]{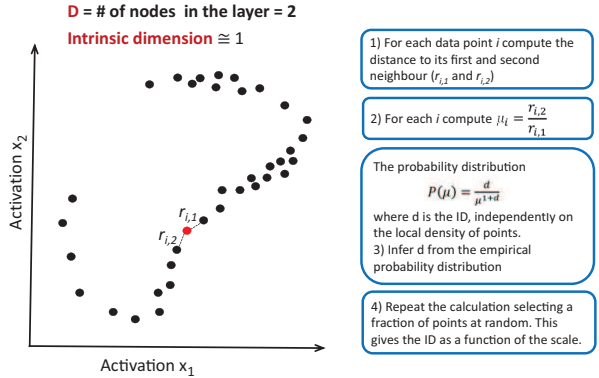}
\caption{The TwoNN estimator derives an estimate of intrinsic dimensionality from the statistics of nearest-neighbour distances.}
\label{fig:twoNN}
\vspace{-.5cm}
\end{wrapfigure}

However, there has not been a direct and systematic characterization of how the intrinsic dimension of data manifolds varies across the layers of CNNs and how it relates to generalization across a wide variety of architectures. We here leverage TwoNN \cite{facco2017estimating}, a recently developed estimator for \emph{global} ID that exploits the fact that nearest-neighbour statistics depend on the ID \cite{LevinaBickel_05} (see Fig. \ref{fig:twoNN} for an illustration). TwoNN can be applied even if the manifold containing the data is curved, topologically complex, and sampled non-uniformly. 
This procedure is not only accurate, but also computationally efficient. In a few seconds on a desktop PC it provides the estimate of the ID of a data set with $O(10^4)$ data, each with  $O(10^5)$ coordinates (for example the activations in an intermediate layer of a CNN), thus making it possible to map out ID across multiple layers and networks. Using this estimator, we investigated the variation of the ID along the layers of a wide range of deep neural networks trained for image classification. Specifically, we addressed the following questions:

\begin{itemize}[noitemsep,topsep=0pt,leftmargin=20pt]
    \item How does the ID change along the layers of CNNs? Do CNNs compress representations into low-dimensional manifolds, or, conversely, seek to expand the dimensionality?
    \item How different is the actual ID from the `linear' dimensionality of a network, i.e., the dimensionality of the linear subspace containing the data-manifold? 
     A substantial mismatch would indicate that the underlying manifolds are curved rather than flat.
    \item How is the ID of a network related to its generalization performance? Can we find empirical signatures of generalization performance in the geometrical structure of the representations? 
\end{itemize}

Our analyses show that data representations in CNNs are embedded in manifolds of low dimensionality, which is typically several orders of magnitude lower than the dimensionality of the embedding space (the number of units in a layer).
In addition, we found that the variation of the ID along the hidden layers of CNNs follows a similar trend across different architectures -- the early layers expand the dimensionality of the representations, followed by a monotonic decrease that brings the ID to reach low values in the final layers. 

Moreover, we observed that, in networks trained to classify images, the ID of the training set in the last hidden layer is an accurate predictor of the network's classification accuracy on the test set -- i.e, the lower the ID in this layer, the better the network capability of correctly classifying the image categories in a test set. Conversely, in the last hidden layer, the ID remains high for a network trained on non predictable data (i.e., with permuted labels), on which the network is forced to memorize rather than generalize.
These geometrical properties of representations in trained neural networks were empirically conserved across multiple architectures, and might point to an operating principle of deep neural networks. 

\section{Estimating the intrinsic dimension of data representations}

Inferring the intrinsic dimension of high-dimensional and sparsely sampled data representations is a challenging statistical problem. To estimate the ID of data-representations in deep networks, we leverage a recently developed global ID-estimator (`TwoNN') that is based on computing the ratio between the distances to the second and first nearest neighbors (NN) of each data point \cite{facco2017estimating} (see Fig. \ref{fig:twoNN}). This allows overcoming the problems related to the curvature of the embedding manifold and to the local variations in the density of the data points, under the weak assumption that the density is constant on the scale of the distance between each point and its second nearest neighbor. 

Formally, let points $x_i$ be uniformly sampled on a manifold with intrinsic dimension $d$ and let $N$ be the total number of points. Let {$r_{i}^{(1)}$} and {$r_{i}^{(2)}$} be the distances of the first and second neighbor of $i$ respectively. Then {$\mu_i \doteq r_{i}^{(2)}/r_{i}^{(1)}, i=1,2,...,N$} follows a Pareto distribution with parameter $d+1$ on $[1, + \infty)$, $ f(\mu_i | d)=d \mu_i^{-(d+1)}$. Taking advantage of this observation, we can formulate the likelihood of vector $\pmb{\mu} \doteq (\mu_1, \mu_2,...,\mu_N)$ as
\begin{equation}
P(\pmb{\mu}|d)=d^N\prod_{i=1}^N \mu_i^{-(d+1)}.
\label{basic}
\end{equation}

At this point $d$ can be easily computed, for instance by maximizing the likelihood, or, following \cite{facco2017estimating}, by employing the empirical cumulate of the distribution of the $\mu$ values to reduce the ID estimation task to a linear regression problem. Indeed, the ID can also be estimated by restricting the product in eq. \ref{basic} to non-intersecting triplets of points, for which independence is strictly satisfied, but, as shown in ref. \cite{facco2017estimating}, in practice this does not significantly affect the estimate.

The ID estimated by this approach is asymptotically correct even for samples harvested from highly non-uniform probability distributions. For a finite number of data points, the estimated values remain very close to the ground truth ID, when this is smaller than $\sim$ 20. 
For larger IDs and finite sample size, the approach moderately underestimates the correct value, especially if the density of data is non-uniform. Therefore, the values reported in the following figures, when larger $\sim$ 20, should be considered as lower bounds. 

For real-world data, the intrinsic dimension always depends on the scale of distances on which the analysis is performed. This implies that the reliability of the dimensionality estimate needs to be assessed by measuring the intrinsic dimension at different scales and by checking whether it is, at least approximately, scale invariant \cite{facco2017estimating}. In our analyses, this test was performed by systematically decimating the dataset, thus gradually reducing its size. The ID was then estimated on the reduced samples, in which the average distance between data points had become progressively larger. 
This allowed estimating the dependence of the ID on the scale.
As explained in \cite{facco2017estimating}, if the ID is well-defined, its estimated value will only depend weakly on the number of data points $N$; in particular it will be not severely affected by the presence of “hubs”, since the decimation procedure would kill them (see Fig. \ref{fig:vgg16r}B).

To test the reliability of our ID estimator on embedding spaces with a dimension comparable to that found in the layers of a deep network, we performed  tests on artificial data of known ID, embedded in a 100,000 dimensional space. The test did not reveal any significant degradation of the  accuracy. Indeed, the ID estimator is sensitive only to the value of the distances between pair of points, and this distance does not depend on the embedding dimension.

For computational efficiency, we analyzed the representations of a subset of layers. We extracted representations at pooling layers after a convolution or a block of consecutive convolutions, and at fully connected layers. In the experiments with ResNets, we extracted the representations after each ResNet block \cite{He2015} and the average pooling before the output.
See \ref{numexp} for details. 

The code to compute the ID estimates with the TwoNN method and to reproduce our experiments is available at this repository: \href{https://github.com/ansuini/IntrinsicDimDeep}{github.com/ansuini/IntrinsicDimDeep}.

\section{Results}

\subsection{The intrinsic dimension exhibits a characteristic shape across several networks}\label{subsec:hunchbacks}

\begin{figure}[htbp!]
\centering
	\includegraphics[clip=true,width=0.9\linewidth]{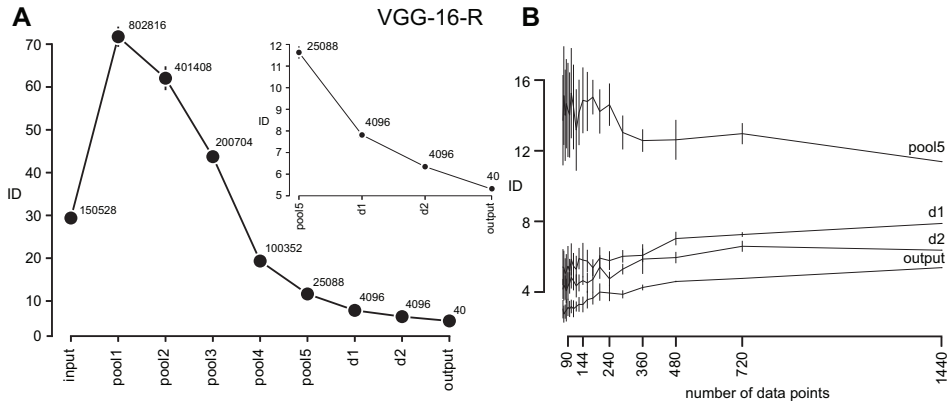}
	\caption{{\bf Modulation of ID across hidden layers of deep convolutional networks} \textbf{A)} ID across layers of VGG-16-R, error bars are the standard deviation of the ID (see \ref{numexp}). Numbers in plot indicate embedding dimensionality of each layer. 
	\textbf{B} Subsampling analysis on VGG-16-R experiment, reported for the same layers as in the inset in \textbf{A} (see \ref{numexp} for details).
	}
\label{fig:vgg16r}
\end{figure}

Our first goal was to empirically characterize the ID of data representations in different layers of deep neural networks. Given a layer $l$ of a DNN, an individual data point (e.g., an image) is mapped onto the set of activations of all the $n_l$ units of the layer, which define a point in a $n_l$-dimensional space. We refer to $n_l$ as the embedding dimension (ED) of the representation in layer $l$. A set of $N$ input samples (e.g., $N$ images) generate, in each layer $l$, a set of $N$ $n_l$-dimensional points. We estimated the dimension of the manifold containing these points using TwoNN.

We first investigated the variation of the ID across the layers of a VGG-16 network \cite{simonyan2014very}, pre-trained on ImageNet \cite{krizhevsky2012imagenet}, and fine-tuned and evaluated on a synthetic data-set of 1440 images \cite{vascon2018characterization}. The dataset consisted of 40 3D objects, each rendered in 36 different views (we left out 6 images for each object as a test set) -- it thus spanned a spectrum of different appearances, but of a small number of underlying geometrical objects. When estimating the ID of data representations on this network (referred to as `VGG-16-R'), we found that the ID first increased in the first pooling layer, before successively and monotonically decreasing across the following layers, reaching very low values in the final hidden layers (Fig. \ref{fig:vgg16r}A). For instance, in the fourth layer of pooling (pool4) of VGG-16-R, ID $\approx 19$ and ED $\approx 10^5$, with $\frac{ID}{ED} \approx 2\times 10^{-4}$. 

One potential concern is whether the number of stimuli is sufficient for the ID-estimate to be robust. To investigate this, we repeated the analysis on subsamples randomly chosen on the data manifold, finding that the estimated IDs were indeed stable across a wide range of sample sizes (Fig. \ref{fig:vgg16r}B). We note that, for the early/intermediate layers, the reported values of the ID are likely a lower bound to the real ID (see discussion in \cite{facco2017estimating}).

\begin{figure}[t!]
	\centering
	\includegraphics[clip=true,width=1\textwidth]{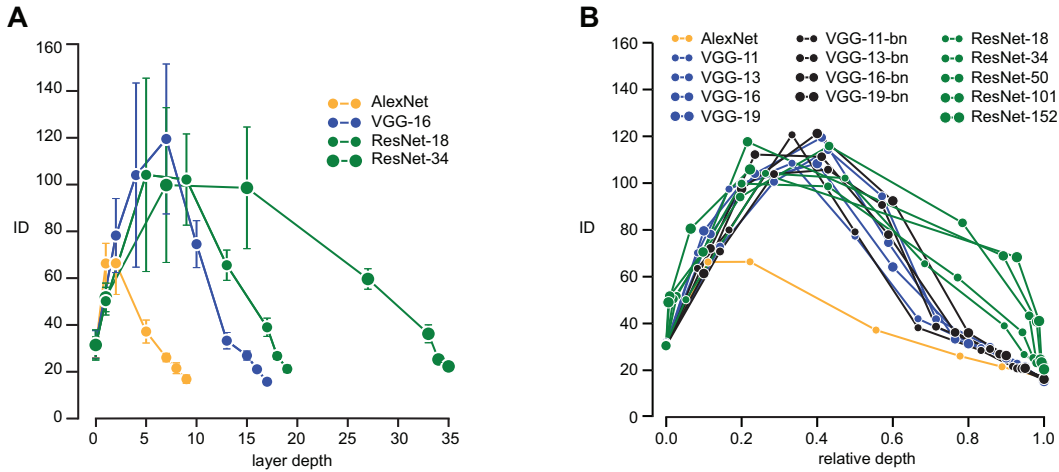} 
	\caption{\textbf{ID of object manifolds across networks. A)} IDs of data representations for 4 networks: each point is the average of the IDs of 7 object manifolds. The error bars are the standard deviations of the ID across the single object's estimates (see \ref{numexp}).  \textbf{B)} The ID as a function of the relative depth in 14 deep convolutional networks spanning different sizes, architectures and training techniques. Despite the wide diversity of these models, the ID profile follows a typical hunchback shape (error bars not shown).
	}
\label{fig:hunchbacks}
\end{figure}

Are the ‘hunchback’ shape of the ID variation across the layers (i.e., the initial steep increase followed by a gradual monotonic decrease), and the overall low values of the ID, specific to this particular network architecture and dataset? To investigate this question, we repeated these analyses on several standard architectures (AlexNet, VGG and ResNet) pre-trained on ImageNet \cite{imagenet_cvpr09}. Specifically, we computed the average ID of the object manifolds corresponding to the 7 biggest ImageNet categories, using ~500 images per category (see section \ref{numexp}). We found both the hunchback-shape and the low IDs to be preserved across all networks (Fig. \ref{fig:hunchbacks}A): the ID initially grew, then reached a peak or a plateau and, finally, progressively decreased towards its final value. As shown in Fig. \ref{fig:suppl_alexnet_categories} for AlexNet, such profile of ID variation across layers was generally consistent across object classes.

The ID in the output layer was the smallest, often assuming a value of the order of ten. Such a low value is to be expected, given that the  ID of the output layer of a network capable of recognizing $N_c$ categories is bound by the condition $N_c \le 2^{ID}$, which implies that each category is associated with a binary representation, and that the output layer optimally encodes this representation. For the $\sim 1000$ categories of ImageNet, this bound  becomes $ID \gtrsim 10$, a value consistent with those observed in all the networks we considered.

Is the relative (rather than the absolute) depth of a layer indicative of the ID? To investigate this,  we plotted ID against relative depth (defined as the absolute depth of the layer divided by the total number of layers, not counting batch normalization layers 
\cite{raghu2017svcca}) of the 14 models belonging to the three classes of networks (Fig. \ref{fig:hunchbacks}B). Remarkably, the ID profiles approximately collapsed onto a common hunchback shape \footnote{with the exception of AlexNet, and a  small network trained on MNIST in a separate analysis, see section \ref{subsec:grad_lum} for details and analysis}, despite considerable variations in the architecture, number of layers, and optimization algorithms. 
For networks belonging to the VGG and ResNet families, the rising portions of the ID profiles substantially overlapped, with the ID reaching similar large peak values (between 100 and 120) in the relative depth range 0.2-0.4. The dependence on relative depth is consistent with the results of \cite{raghu2017svcca}, where it was observed that similarity between layers depended on relative depth.

Notably, in all networks the ID eventually converged to small values in the last hidden layer. These results suggest that state-of-the-art deep neural networks -- after an initial increase in ID -- perform a progressive  dimensionality reduction of the input feature vectors, a result with is consistent with the information-theoretical analysis in \cite{AchilleSoatto_18}.
Based on previous findings about the evolution of the ID in the last hidden layer during training \cite{Ma_2018}, one could speculate that this progressive, gradual reduction of dimensionality of data-manifolds is a feature of deep neural networks which allows them to generalize well. In the following, we will investigate this idea further by showing that the ID of the last hidden layers predicts generalization performance, and by showing that these properties cannot be found in networks with random weights or trained on non predictable data.

\subsection{The intrinsic dimension of the last hidden layer 
predicts classification performance} \label{subsec:pretrained}

\begin{wrapfigure}{u}{.5\textwidth}
\includegraphics[width=.5\textwidth]{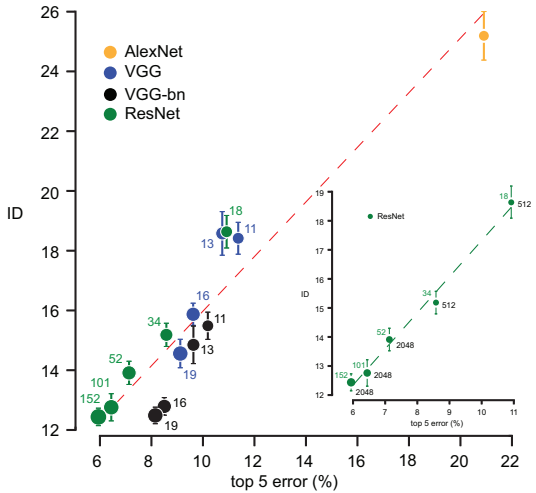}
	\caption{\textbf{ID of the last hidden layer predicts performance.} The ID of data representations (training set) predicts the top 5-score performance on the test set. \textbf{Inset} Detail for the ResNet class.}
\label{fig:lasthidden}
\end{wrapfigure}

Although the hunchback shape was preserved across networks, the IDs in the last hidden layers were not exactly the same for all the networks. To better resolve such differences, we computed the ID in the last hidden layer of each network using a much larger pool of images of the training set ($\sim 2,000$), sampled from all ImageNet categories (see section \ref{numexp}). This revealed a spread of ID values, ranging between $\approx 12$ (for ResNet152) and $\approx 25$ (for AlexNet, Fig. \ref{fig:lasthidden}). These differences may appear small, compared to the much larger size of the embedding space in the last hidden layer (where the ED was between $1$ and $2$ orders of magnitude larger than the ID (range $=[512-4096]$).  However,  the ID in the last hidden layer on the training set was indeed a strong predictor of the performance of the network on the test set, as measured by top 5-score 
(Fig. \ref{fig:lasthidden}, Pearson correlation coefficient $r = 0.94$). A tight correlation was found not only across the full set of networks, but also within each class of architectures, when such comparison was possible -- i.e., in the classes of the VGG with and without batch normalization and ResNets ($r=0.99$ in the latter case, see inset in Fig. \ref{fig:lasthidden}).

Overall, this analysis suggests that the ID in the last hidden layer can be used as a proxy for the generalization ability of a network. Importantly, this proxy can be measured without estimating the performance on an external validation set. 

\subsection{Data representations lie on curved manifolds}\label{subsec:curved_manifolds}

The strength of the TwoNN method lies in its ability to infer the ID of data representations, even if they lie on curved manifolds.
This raises the question of whether our observations (low IDs, hunchback shapes,  correlation with test-error)  reflect the fact that data points live on low-dimensional, yet highly curved manifolds, or, simply, in low-dimensional, but largely flat (linear) subspaces.

\begin{figure}[htbp!]
\includegraphics[clip=true,width=1\textwidth]{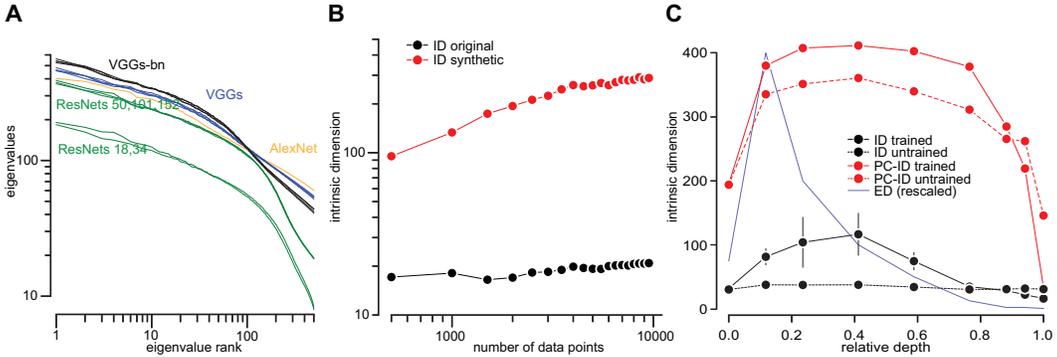} 
\caption{\textbf{Evidence that data-representations are on curved manifolds} \textbf{A)} Variance spectra of last hidden layer  do not show a clear gap. \textbf{B)} ID in the last hidden layer of VGG-16 (black), compared with the ID of a synthetic Gaussian dataset with the same size and second-order correlations structure (red).
\textbf{C)} The ID and the PC-ID along the layers of VGG-16 for a trained network and an untrained, randomly initialized network. The ED, rescaled to reach the maximum at 400, is shown in blue.
}

\label{fig:curvature}
\end{figure}

To test this, we performed linear dimensionality reduction (principal component analysis, PCA) on the normalized covariance matrix (i.e., the matrix of correlation coefficients -- using the raw covariance resulted in qualitatively similar results) for each layer and network. We did not find a clear gap in the eigenvalue spectrum (Fig. \ref{fig:curvature}A), a  result that is qualitatively consistent with that obtained for stimulus-representations in primary visual cortex \cite{stringer2018high}. The absence of a gap in the spectrum, with the magnitude of the eigenvalues smoothly decreasing as a function of their rank, is, by itself, an indication that the data manifolds are not linear.
Nevertheless, we defined an `ad-hoc' estimate of dimensionality by computing the number of components that should be included to describe $90 \%$ of the variance in the data. In what follows, we call this number PC-ID. We found PC-ID to be about one or two orders of magnitude larger than the value of the ID computed with TwoNN. For example, the PC-ID in the last hidden layer of VGG-16 was $\approx 200$ (Fig. \ref{fig:curvature}C, solid red line), while the ID estimated with TwoNN was $\approx 18$ (solid black line).

The discrepancy between the ID estimated with TwoNN and with PCA points to the existence of strong non-linearities in the correlations between the data, which are not captured by the covariance matrix. To verify that this was indeed the case (and, e.g., not a consequence of estimation bias), we used TwoNN to compare the ID of the last hidden layer of VGG-16 with the ID of a synthetic Gaussian dataset with the same second-order moments. The ID of the original dataset was low and stable as a function of the size $N$ of the data sample used to estimate it (Fig. \ref{fig:curvature}B, black curve; similar subsampling analysis as previously shown in Fig. \ref{fig:vgg16r}B). In contrast, the ID of the synthetic dataset was two orders of magnitude larger, and grew with $N$ (Fig. \ref{fig:curvature}B, red curve),  as expected in the case of an ill-defined estimator \cite{facco2017estimating}. 

We also computed the PC-ID of the object manifolds across the layers of VGG-16 on randomly initialized networks, and we found that its profile was qualitatively the same as in trained networks (compare solid and dashed red curves in Fig. \ref{fig:curvature}C). By contrast, when the same comparison was performed on the ID (as computed using TwoNN), the trends obtained on random weights  (dashed black curve) and after training the network (solid black curve) were very different. While the latter showed the hunchback profile (same as in Fig. \ref{fig:hunchbacks}), the former was remarkably flat.
This behaviour can be explained by observing that the ID of the input is very low (see section \ref{subsec:grad_lum} for a discussion of this point). For random weights, each layer effectively performs an orthogonal transformation, thus preserving such low ID across layers. Importantly, the hunchback profile observed for the ID in trained networks (Figs \ref{fig:vgg16r}A, \ref{fig:hunchbacks}A,B) is a genuine result of training, which does not merely reflect the initial expansion of the ED from the input to the first hidden layers, as shown by the fact that, in VGG-16, the ID kept growing after that the ED had already started to substantially decline (compare the solid black and blue curves in Fig. \ref{fig:curvature}C).

The analysis shown in Fig. \ref{fig:curvature}C also indicates that intermediate layers and the last hidden layer undergo opposite trends, as the result of training (compare the solid and dashed black curves): while the ID of the last hidden layer is reduced with respect to its initial value [consistently with what reported in \cite{Ma_2018}], the ID of intermediate layers increases by a large amount. This prompted us to run an exploratory analysis to monitor the ID evolution during training in a VGG-16 network trained with CIFAR-10. We observed a behavior that was consistent with that already reported in Fig. \ref{fig:curvature}C: a substantial increase of the ID in the initial/intermediate layers, and a decrease in the last layers (Fig. \ref{fig:suppl_dynamics}A, black vs. orange curve). Interestingly, a closer inspection of the dynamics in the last hidden layer revealed a non-monotonic variation of the ID (see Fig. \ref{fig:suppl_dynamics}B,C). Here, after an initial drop, the ID slowly increased, but, differently from \cite{Ma_2018}, without resulting in substantial overfitting. Thus, the evolution of the ID during learning appears to be not strictly monotonic and its trend likely depends on the specific architecture and dataset, calling for further investigation.

\subsection{The initial increase in intrinsic dimension can arise from irrelevant features}\label{subsec:grad_lum}

\begin{figure}[htbp!]
	\centering
	\includegraphics[clip=true,width=0.8\textwidth]{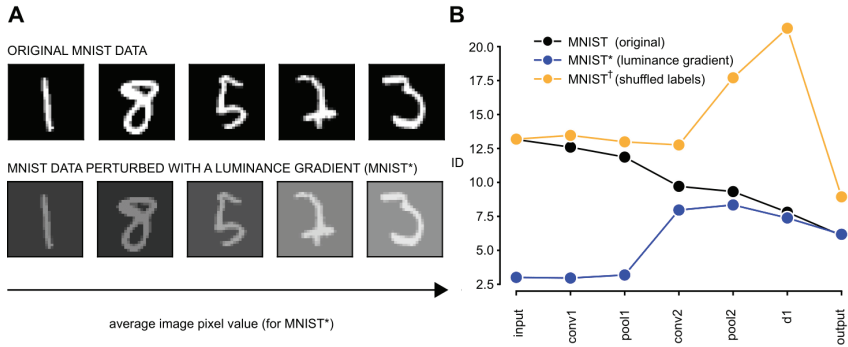} 
	\caption{\textbf{A} The addition of a luminance gradient across the images of the MNIST dataset results in a stretching of the image manifold along a straight line in the input space of the pixel representation. \textbf{B} Change of the ID along all the layers of the MNIST network, as obtained in three different experiments: 1) with the original MNIST dataset (black curve) 2) with the luminance-perturbed MNIST$^\star$ dataset (blue curve) and 3) with the MNIST$^\dagger$, in which the label of the MNIST images where randomly shuffled (red curve).} 
	\label{fig:mnist}
\end{figure}

We generally found the ID to increase in the initial layers.
However, this was not observed for a small network trained on the MNIST data-set (Fig. \ref{fig:mnist}B, black curve) and was also less pronounced for AlexNet (Fig. \ref{fig:hunchbacks}A, orange curve). A mechanism underlying the initial ID rise could be the fact that the input is dominated by features that are irrelevant for predicting the output, but are highly correlated between each other. To validate this hypothesis, we generated a modified MNIST dataset (referred to as MNIST$^\star$) by adding a luminance perturbation that was constant for all pixels within an image, but random across the various images (Fig. \ref{fig:mnist}A). Given an image $i$ with pixels of  $x_i \in \mathcal{R}^N$ (where $N$ is the number of pixels), we added shared random perturbations,  $x_i \rightarrow x_i^\star = x_i + \lambda \xi_i$ where $\lambda$ is a positive parameter and $\xi_i$ are i.i.d. uniformly distributed random variables in the range $[0,1]$. This perturbation has the effect of stretching the dataset along a specific direction in the input space (the vector $[1,1,\dots,1]$) thus reducing the ID of the data manifold in the input layer. Indeed, with $\lambda = 100$, the ID of the input representation dropped from $\approx 13$ (its original value) to $\approx 3$.

The network trained on MNIST$^\star$ was still able to generalize (accuracy $\approx 98 \%$). However, the variation of the ID (blue curve in Fig. \ref{fig:mnist}B) now showed a hunchback shape reminiscent of that already observed in Figs \ref{fig:vgg16r}A and \ref{fig:hunchbacks}A,B for large architectures. 
This suggests that the growth of the ID in the first hidden layers of a deep network is determined by the presence in the input data of low-level features that carry no information about the correct labeling -- for instance, in the case of images, gradients of luminance or contrast. 
One can speculate that, in a trained deep network, the first layers prune the irrelevant features, formatting the representation for the more advanced processing carried out by the last layers \cite{AchilleSoatto_18}. The initial increase of the dimensionality of the data manifold could be the signature of such pruning. This notion is consistent with recent evidence gathered in the field of visual neuroscience, where the pruning of low-level confounding features, such as luminance, has been demonstrated along the progression of visual cortical areas that, in the rat brain, 
are thought to support shape processing and object recognition \cite{tafazoli2017emergence}.

\subsection{A network trained on random labels does not show the characteristic hunchback profile of ID variation}\label{subsec:random_labels} 

In untrained networks the ID profile is largely flat (Fig. \ref{fig:curvature}C). Are there other circumstances in which the ID profile deviates from the typical hunchback shape of Figs \ref{fig:vgg16r}A and \ref{fig:hunchbacks}A,B, with IDs that do not decrease progressively towards the output? It turns out that this is the case when generalization is impossible by construction, as we verified by randomly shuffling the labels on MNIST (we refer to the shuffled data as MNIST$^\dagger$). 

It has been shown \cite{Zhang2017} that deep networks can perfectly fit the training set on randomly labelled data, while necessarily achieving chance level performance on the test set. As a result, when we trained the same network as in section \ref{subsec:grad_lum} on MNIST$^\dagger$, we achieved a training error of zero. However, we found that the network had an ID profile which did not decrease monotonically (orange curve in Fig. \ref{fig:mnist}B) -- in contrast to the same network trained with the original dataset (black curve). Instead, it grew considerably in the second half of the network, almost saturating the upper bound, which is set by the ED, in the output layer. 
This suggests that the reduction of the dimensionality of data manifolds in the last layers of a trained network reflects the process of learning on a generalizable dataset. By contrast, overfitting noisy labels leads to an expansion of the dimensionality, as already reported in \cite{Ma_2018}. As suggested in that study, this indicates that a network trained on inconsistent data can be recognized \emph{without} estimating its performance on a test set, but by simply looking at whether the ID increases substantially across its final layers.

\section{Conclusions and Discussion}

Convolutional neural networks, as well as their biological counterparts, such as the visual system of primates \cite{dicarlo2012does} and other species \cite{tafazoli2017emergence, matteucci2019nonlinear}, transform the input images across a progression of processing stages, eventually providing an explicit (i.e. transformation-tolerant) representation of visual objects in the output layer. Leading theories in the field of visual neuroscience postulate that such re-formatting gradually untangles and flattens the manifolds produced by the different images 
within the representational space defined by the activity of all the neurons (or units) in a layer \cite{dicarlo2012does, dicarlo2007untangling, olshausen2004sparse}. This suggests that the dimensionality of the object manifolds may progressively decrease along the layers of a deep network, and that such a decrease may be at the root of the high classification accuracy achieved by deep networks. Although previous theoretical and empirical studies have provided support to this hypothesis using small/simple network architectures or focusing on single layers of large networks \cite{huang2018mechanisms, yu2018curvature, Ma_2018,  basri2016efficient, chung2017manifolds, brahma2016deep},our study is the first to investigate systematically how the dimensionality of individual object manifolds - or mixtures of object manifolds - vary in large, state-of-the-art CNNs used for image classification.

\begin{wrapfigure}{r}{.45\textwidth}
\includegraphics[width=.45\textwidth]{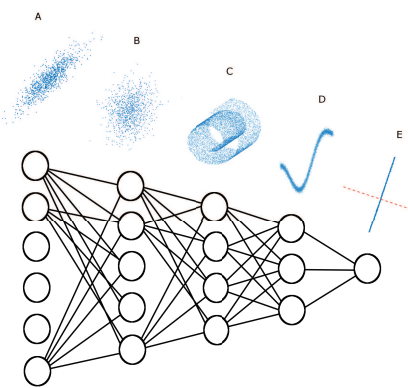}
\caption{\textbf{A}. Input layer. The intrinsic dimensionality of the data can assume low values due to the presence of irrelevant features uncorrelated with the ground truth. \textbf{B}. The first hidden layers pre-process the data raising its intrinsic dimension. \textbf{C,D}. The representation is squeezed onto manifolds of progressively lower intrinsic dimension. These manifolds are typically not hyperplanes. \textbf{D}. In the last hidden layer (\textbf{D}) the ID shows a remarkable correlation with the performance in trained networks. \textbf{E}. The output layer.}
\label{fig:wrap_up}
\vspace{-.3cm}
\end{wrapfigure}

Our results can be summarized by making reference to the cartoon shown in Fig. \ref{fig:wrap_up}. We found that the ID in the initial layer of a network is low. As shown in Fig. \ref{fig:mnist}, this can be explained by the existence of gradients of correlated low-level visual features (e.g., luminance, contrast, etc.) across the image set  \cite{simoncelli2001natural}, resulting in a stretching of image representations along a few, main variation axes within the input space (see Fig. \ref{fig:wrap_up}A). Early layers of DNNs appear to 
get rid of these  correlations, which are irrelevant for the classification task, thus leading to an increase of the ID of the object manifolds (Fig. \ref{fig:vgg16r}A and \ref{fig:hunchbacks}A,B). As illustrated in Fig. \ref{fig:wrap_up}B, this can be thought as a sort of whitening of the input data. Such initial dimensionality-expansion is also thought to be performed in the visual system \cite{olshausen2004sparse, simoncelli2001natural}, and is consistent with a recent characterization of the dimensionality of image representations in  primary visual cortex \cite{stringer2018high} and with the pruning of low-level information performed by high-order visual cortical areas \cite{tafazoli2017emergence}.

After this initial expansion, the representation is squeezed into manifolds of progressively lower ID (Figs \ref{fig:vgg16r}, \ref{fig:hunchbacks}A,B), as  graphically illustrated in Fig. \ref{fig:wrap_up}C,D). This phenomenon has been already observed by \cite{basri2016efficient} and \cite{huang2018mechanisms} on simplified datasets and architectures, by \cite{yu2018curvature} in the final, fully connected layers of AlexNet, and by \cite{Ma_2018} in the last hidden layers of two different DNNs, where the ID evolution was tracked during training. 
We here demonstrate that this progressive reduction of the dimension of data manifolds is a general behavior and a key signature of every CNN we tested -- both small toy models (Fig. \ref{fig:mnist}B) and large state-of-the-art networks  (Fig. \ref{fig:hunchbacks}A,B). More importantly, our experiments show that the extent to which a deep network is able to compress the dimensionality of data representations in the last hidden layer is a key predictor of its ability to generalize well to unseen data (Fig. \ref{fig:lasthidden}) -- a finding that is consistent with the inverse relationship between ID and accuracy reported by \cite{Ma_2018}, although our pilot tests suggest that the ID, after a large, initial drop, can slightly increase during training without producing overfitting (Fig. \ref{fig:suppl_dynamics}). From a theoretical standpoint, this result is broadly consistent with recent studies linking the classification capacity of data manifolds by perceptrons to their geometrical properties \cite{chung2017manifolds, chung2016linear}.
Our findings also resonate with the compression of the information about the input data during the final phase of training of deep networks \cite{shwartz2017opening}, which is progressively larger as a function of the layer’s depth, thus displaying a trend that is reminiscent of the one observed for the ID in our study. 

Finally, our experiments also show that the ID values are lower than those identified using PCA, or on `linearized' data, which is an indication that the data lies on curved manifolds. In addition, ID measures from PCA did not qualitatively distinguish between trained and randomly initialized networks (Fig. \ref{fig:curvature}C). 
This conclusion is at odds with the unfolding of data manifolds reported by \cite{brahma2016deep} across the layers of a small network tested with simple datasets. It also suggests a slight twist on theories about transformations in the visual system \cite{dicarlo2012does, dicarlo2007untangling} -- it indicates that a flattening of data manifolds may not be a general computational goal that deep networks strive to achieve: progressive reduction of the ID, rather than gradual flattening, seems to be the key to achieving linearly separable representations.
 
To conclude, we hope that data-driven, empirical approaches to investigate deep neural networks, like the one implemented in our study, will provide intuitions and constraints, which will ultimately inspire and enable the development of theoretical explanations of their computational capabilities. 

\clearpage

\subsubsection*{Acknowledgments}

We thank Eis Annavini for providing the custom dataset described in \ref{custom_dataset}; Artur Speiser and Jan-Matthis Lückmann for a careful proofreading of the manuscript. We warmly thank Elena Facco for her valuable help in the early phases of this project. We also thank Naftali Tishby, Riccardo Zecchina, Matteo Marsili, Tim Kietzmann, Florent Krzkala, Lenka Zdeborova, Fabio Anselmi, Luca Bortolussi, Jim DiCarlo and SueYeon Chung for useful discussions and suggestions, and the anonymous referees for their useful and constructive comments.

This work was supported by a European Research Council (ERC) Consolidator Grant, 616803-LEARN2SEE (D.Z.). JHM is funded by the German Research Foundation (DFG) through  SFB 1233 (276693517), SFB 1089 and SPP 2041, the German Federal Ministry of Education and Research (BMBF, project `ADMIMEM', FKZ 01IS18052 A-D), and the Human Frontier Science Program (RGY0076/2018).



\clearpage

\appendix

\section{Appendix}

\subsection{Details of numerical experiments}\label{numexp}

All our experiments were performed in PyTorch \cite{paszke2017automatic} (version 1.0) on a Linux workstation with 64GB of RAM and a GeForce GTX 1080 Ti NVIDIA graphic card. 
The code to compute the ID estimates with the TwoNN method and to reproduce our experiments is available at this repository: \href{https://github.com/ansuini/IntrinsicDimDeep}{github.com/ansuini/IntrinsicDimDeep}. The data is downloadable at this link: \href{https://figshare.com/s/8a039f58c7b84a215b6d}{https://figshare.com/s/8a039f58c7b84a215b6d}.

\subsubsection{Datasets}

\paragraph{Custom dataset}\label{custom_dataset}

A dataset of 1400 images developed for a neurophysiological study \cite{vascon2018characterization}. The dataset consisted of 40 three-dimensional (3D), computer graphics models of both natural and man-made objects, each rendered in 36 different views, obtained by combining in-plane and in-depth rotations of the 3D models with horizontal translations and size variations. As a result, the image set encompassed a spectrum of object identities, poses and low-level features (e.g., luminance, contrast, position, size, aspect ratio, etc.), but without reaching the size, complexity and variety of shapes and identity-preserving transformations that are typical of naturalistic image sets, such as ImageNet.

\subsubsection{Architectures}\label{architectures}

We describe the architectures used in order of appearance in the main text.
\paragraph{VGG-16-R}

We removed the last hidden layers (the last convolutional and all the dense layers) of a VGG-16 network \cite{simonyan2014very} pre-trained on ImageNet \cite{krizhevsky2012imagenet} and substituted it with randomly initialized layers of the same size except for the last hidden layer, in order to match the correct number of categories (40) of the custom dataset described in \ref{custom_dataset}.
We then fine-tuned it on the $\simeq 85 \%$ of the data.
More specifically we used $30$ images for each category as training set and we tested on the remaining $6$ images for each category. We called this network VGG-16-R (where R stands for restricted, with reference to this small dataset). For the fine-tuning, we used a SGD with momentum $0.9$, and a learning rate of $10^{-4}$ in the last convolutional layer and of $10^{-3}$ in the dense layers. The other layers were kept frozen. The generalization performance after 15 epochs was $\approx 88 \%$ accuracy on the test set.

\paragraph{Standard architectures pre-trained on ImageNet}

We instantiated fourteen pre-trained networks that are representative of the state-of-the-art models used in visual object recognition and image understanding: AlexNet \cite{krizhevsky2012imagenet}, eight models belonging to the VGG class (11,13,16 and 19 with and without batch normalization) \cite{simonyan2014very}), and five models belonging to the ResNet class (18,34,52,101,152) \cite{He2015}. All these models are available for download in Pytorch \cite{paszke2017automatic} at \href{https://pytorch.org/docs/stable/torchvision/models.html}{torchvision/models.html}.

\paragraph{Small convolutional network for the experiments on the MNIST dataset}

We trained a small convolutional network on the MNIST dataset \cite{mnist}. The sequence of layers is: a convolutional layer with 1 input channel, 32 output channels and a kernel size of 3; a max pooling layer of kernel size 2; a convolutional layer with 32 input channels, 64 output channels and a kernel size of 3; a max pooling layer of kernel size 2; a fully connected layer with 1600 inputs and 128 outputs; a fully connected layer with 128 input and 10 output units; a  softmax. We used ReLU non-linearity after each convolutional, pooling and fully connected layer. The stride is always set to zero. The network has been trained for $200$ epochs with a small learning rate ($lr=0.0004$) and zero momentum on the original dataset, for $5000$ epochs $lr=0.0001$ and momentum $0.9$ on MNIST$^\star$ and for $500$ epochs $lr=0.0005$ and momentum $0.9$ on MNIST$^\dagger$.

\paragraph{VGG-16 adapted for CIFAR-10}

We used the VGG-16 model adapted for CIFAR-10 available at \href{https://github.com/kuangliu/pytorch-cifar}{github.com/kuangliu/pytorch-cifar} in two series of experiments. In our experiment on the dependency of the ID (in the last hidden layer) on random initialization (results are in Sec. \ref{further_results}) we trained the network for $300$ epochs starting with a learning rate of $0.1$ and reducing it by a factor $0.1$ after every $100$ epochs. $L_2$ regularization was applied with a weight decay set at $5\times 10^{-4}$. In two experiments on dynamics we used the same configuration (see Fig. \ref{fig:suppl_dynamics}A) and one with a smaller learning rate $lr=0.005$ (Fig. \ref{fig:suppl_dynamics}B,C), in order to elucidate better the early phases of the dynamics.

\paragraph{Checkpoints}

In each experiment we defined architecture-specific checkpoints from where to extract and analyze the representations. The only exception was in the case of the network used for MNIST, where we extracted the representations and performed the analysis in all the layers, in the experiments described in sections \ref{subsec:grad_lum}, \ref{subsec:random_labels}. As a general rule, we always extracted representations at pooling layers after a convolution or a block of consecutive convolutions, and at fully connected layers. In the experiments with ResNets, we extracted the representations after each ResNet block \cite{He2015} and the average pooling before the output. Depending on the computational demands of our experiments, we extracted and analyzed data samples of different sizes, we describe this in the following section \ref{id_estimation}.

\subsubsection{Estimating intrinsic dimension}\label{id_estimation}

\paragraph{Experiments with the custom dataset}
In this experiment (Fig. \ref{fig:vgg16r}A,B), we fine-tuned the last layers of a VGG-16 network pre-trained on ImageNet using the $\approx 80\%$ of the 1440 images of the dataset in \cite{vascon2018characterization} (30 images for each category in the training set, the remaining 6 images for each category as test set). The whole dataset was used to estimate the ID of the representations across the layers of the network. The values of the ID reported in our analysis are the averages resulting from randomly sampling 20 times the $90\%$ of the activations at each checkpoint layer. The error bars are the standard deviations across these estimates.
In the decimation analysis (Fig. \ref{fig:vgg16r}B) we proceeded as described in \cite{facco2017estimating}. After a random shuffling, we splitted the dataset $X$ in a $k$-fold way, with $k$ ranging from 20 to 1. The $k$-fold splits yielded $k$ ID estimates at each layer on roughly $N/k$ of the data. The $k$ ID estimates were then averaged and the standard deviation were computed.

\paragraph{Experiments with ImageNet}

In the experiments with the pre-trained state-of-the-art networks, we performed two kinds of analysis. In the first one (Fig. \ref{fig:hunchbacks}A,B), we sampled randomly 500 images from each of the 7 most populated ImageNet categories: ``koalas'', ``shih-tzu'', ``rhodesian'', ``yorkshire'', ``vizsla'', ``setter'', ``butterfly''. These 7 sets were kept fixed in all the subsequent analysis. Let us call $X_i$ the $i$-th set.
We then estimated the ID of the resulting object manifolds across the layers of the networks, independently for each category. 
For each $i$, we randomly subsampled from the representations of $X_i$ at each checkpoint layer the $90\%$ of the data (450 data points) for 5 times and we computed their IDs. We then averaged these 7 values obtaining a category-specific estimate of the ID at each layer. 
We finally averaged the IDs obtained for the 7 categories and computed their standard deviations.
In the second analysis (Fig. \ref{fig:lasthidden}A,B), we randomly sampled $2000$ images for $5$ times from the ImageNet training set. Let us call $X_i$ the $i$-th of these samples. We computed their representations $R^{\text{last hidden}}_i$ in the last hidden layer, then we randomly subsampled, in each $R^{\text{last hidden}}_i$, the $90\%$ of the data (consisting of 1800 data points) for 20 times and we computed their IDs.
We then pooled together all these 100 ID estimates, computed their average and their standard deviation: these are respectively our final ID estimate and its error. Notice that, in this case, the ID estimates refer to random mixtures of all possible object categories of ImageNet.

\paragraph{Experiments with MNIST}

In these experiments (Fig. \ref{fig:mnist}B, black line), we randomly sampled a set of 2000 images from the test set; this set - called $X$ in the following - was kept fixed. We extracted the activations $R^l$ at each layer $l$ of the trained network described in \ref{architectures}. For $l=0$ the representations are the original images.
For each layer $l$ we randomly subsampled from $R^l$ the $90\%$ of the data (1800 data points) for 50 times and we computed their IDs. We then averaged these ID values and computed their standard deviation: these are respectively our final ID estimate and its error.

\paragraph{Experiments with VGG-16 adapted for CIFAR-10}

In these experiments (Fig. \ref{fig:suppl_dynamics}) we randomly sampled 500 images from each of the 10 CIFAR-10 classes; this set was kept fixed. Then we proceeded similarly to the MNIST case described above.

\clearpage

\subsubsection{Further results}\label{further_results}

\paragraph{Variability of the ID across object categories}

\begin{wrapfigure}{r}{.4\textwidth}
\includegraphics[width=.4\textwidth]{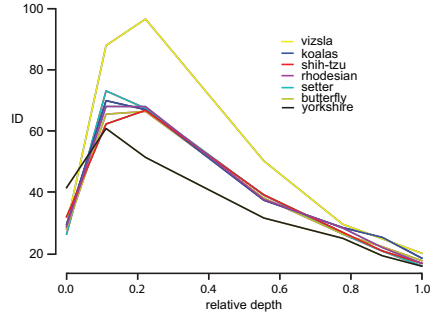}
\caption{The ID variation across layers is generally consistent across object classes. This figure refers to AlexNet in Fig. \ref{fig:hunchbacks}, where the average across classes is reported.}
\label{fig:suppl_alexnet_categories}
\vspace{-.5cm}
\end{wrapfigure}

In section \ref{subsec:hunchbacks}, we showed that, across layers, the ID displays a typical ‘hunchback’ profile. In particular, the average of seven class-specific ID profiles is shown in Fig. \ref{fig:hunchbacks}A,B and their standard deviation is reported in Fig. \ref{fig:hunchbacks}A. To give a better intuition of how consistent this trend was across image categories, the ID profile of the seven individual object classes across the layers of AlexNet is reported in Fig. \ref{fig:suppl_alexnet_categories}.

\paragraph{Variability of the ID across random initializations}

In section \ref{subsec:pretrained}, we showed a correlation between ID in the last hidden layer and test accuracy across a wide variety of architectures. Based on this result, one could wonder if this correlation is also present within a specific model retrained with different random initializations of the weights. To address this question, we estimated the variability of the ID in the last hidden layer of a VGG-16 adapted for CIFAR-10 across 50 different trainings, finding no correlation with accuracy (r=-0.003), likely because of the little variation in accuracy produced by different random weight initializations. This suggests that differences in
accuracy across well-trained networks (see Fig. \ref{fig:lasthidden}) are mostly due to differences in the architecture. 

\paragraph{Dynamics}

\begin{figure}[htbp!]
	\centering
	\includegraphics[clip=true,width=1.0\textwidth]{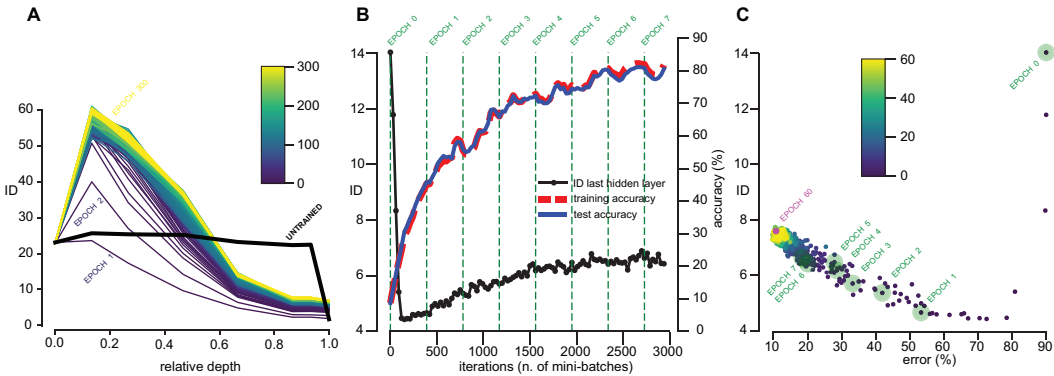} 
	\caption{\textbf{Dynamics of the ID on a VGG-16 trained on CIFAR-10.} \textbf{A}  Dynamics of the ‘hunchback’ shape for a typical training history. The black thick line refers to the untrained network. The color codes for the training epochs. \textbf{B} Training and test accuracy (dashed red and blue curves respectively) and ID in the last hidden layer in the early phases. The dynamics of the ID is non-monotonic; at the same time there is no substantial overfitting. \textbf{C} ID vs. test error (same data as in \textbf{B}).}
	\label{fig:suppl_dynamics}
\end{figure}

In section \ref{subsec:curved_manifolds}, we observed that, during training, the ID of intermediate layers and the last hidden layer undergo opposite trends. To explore more carefully this finding, we monitored the evolution of the ID profile during training of a VGG-16 network with CIFAR-10. We found qualitative confirmation of the observations reported in Fig. \ref{fig:curvature}C, including a flat ID profile in the untrained network (see Fig. \ref{fig:suppl_dynamics}A, black thick curve). Moreover, when we inspected more closely the early phase of the dynamics (by slowing down the learning rate and augmenting the ‘temporal’ resolution of our observations), we found that in the last hidden layer the evolution of the ID is non-monotonic, even in presence of negligeable overfitting (see Fig. \ref{fig:suppl_dynamics}B,C). Differently from what reported by \cite{Ma_2018}, this suggests that the ID in the last hidden layer of a deep network is not always a reliable predictor of the onset of overfit, and whether this is the case may depend on the specific architectures and data used.

\end{document}